\pdfoutput=1

\documentclass[11pt]{article}

\usepackage[]{acl}

\usepackage{times}
\usepackage{mathtools}
\usepackage{latexsym}
\usepackage{amsfonts}
\usepackage{times}
\usepackage{latexsym}
\usepackage{color,soul}
\usepackage{graphicx}
\usepackage{amsmath}
\usepackage{bm}
\usepackage{booktabs}
\usepackage{colortbl}
\usepackage{cleveref}
\usepackage{multirow}
\usepackage{graphicx}
\usepackage{arydshln}
\usepackage{algorithm}
\usepackage{algpseudocode}

\usepackage[T1]{fontenc}

\usepackage[utf8]{inputenc}

\usepackage{microtype}

\usepackage{inconsolata}

\usepackage{graphicx}

\usepackage{algorithm}
\usepackage{algpseudocode}
\usepackage{multirow}
\usepackage{amsmath} 

\usepackage{pgfplots}
\pgfplotsset{compat=1.16}
\usepackage{subcaption}
\usepackage{soul}

\newcommand{\approach}{\textsc{Della}}
\newcommand{\dare}{\textsc{Dare}}
\newcommand{\ties}{\textsc{Ties}}
\newcommand{\ta}{\textsc{TA}}
\newcommand{\magp}{\textsc{MagPrune}}
\newcommand{\topk}{\textsc{TopK}}
\newcommand{\random}{\textsc{Random}}
\newcommand{\nodrop}{\textsc{NoDrop}}
\newcommand{\wanda}{\textsc{Wanda}}

\definecolor{bblue}{HTML}{4F81BD}
\definecolor{rred}{HTML}{C0504D}
\definecolor{ggreen}{HTML}{9BBB59}
\definecolor{ppurple}{HTML}{9F4C7C}
\title{DELLA-Merging: Reducing Interference in Model Merging through Magnitude-Based Sampling}

\author{Pala Tej Deep$^{1}$, Rishabh Bhardwaj$^1$, Soujanya Poria$^1$ \\\\
$^1$ Singapore University of Technology and Design
}

\begin{document}
\maketitle
\begin{abstract}

With the proliferation of domain-specific models, model merging has emerged as a set of techniques that combine the capabilities of multiple models into one that can multitask without the cost of additional training. In this paper, we propose a new model merging technique, \textbf{D}rop and r\textbf{E}sca\textbf{L}e via samp\textbf{L}ing with m\textbf{A}gnitude (\approach{}-Merging), that employs a novel pruning technique, \magp{}, which shows significant advantages over \dare{} and \ties{}. \magp{} first ranks the parameters in order of their magnitude and assigns higher dropout probabilities ($p$) to parameters with lower ranks corresponding to lower magnitudes. To approximate the original embeddings, \magp{} employs a rescaling operation on the parameters that survive the random dropping by $1/(1-p)$. On three different expert models considered for merging (LM, Math, Code) and corresponding benchmark datasets (AlpacaEval, GSM8K, MBPP), \approach{} shows an average improvement of 2.4 points over baseline methods employing delta parameter pruning (an improvement of 3.6 points over \ties{}, 1.2 points over \dare{}), and 11.1 points over the no-pruning baseline (\ta). We release the source code at: \url{https://github.com/declare-lab/della}.

\end{abstract}

\section{Introduction}
Interactive systems based on general-purpose LLMs have become widely popular due to their impressive instruction-following capabilities \cite{openai2023gpt4}. Furthermore, tuning these models on downstream tasks has been shown to transform them into domain experts \cite{codeLlama, luo2023wizardmath}.

Maintaining separate fine-tuned models for each task presents several limitations, such as a significantly higher memory footprint and the inability to leverage information across tasks, which could enhance both in-domain and out-of-domain performance. As a result, merging different homologous models (models fine-tuned from the same backbone) is gaining traction for its cost-effectiveness, knowledge sharing, and space efficiency \cite{ties, dare}. The homologous models differ from each other in terms of \textit{delta} parameters, i.e., the difference between the fine-tuned model and backbone model parameters.

In this paper, we introduce a novel approach for merging homologous models, termed \textbf{D}rop and r\textbf{E}sca\textbf{L}e via samp\textbf{L}ing with m\textbf{A}gnitude (\approach{}). This approach consists of three steps: (Step-1) involves delta parameter drops to reduce interference among model parameters. We propose \textbf{\magp{}}, a novel pruning method that samples delta parameters based on their magnitudes; (Step-2) further reduces interference through sign-based delta parameter selection; and (Step-3) fuses the selected delta parameters.

\begin{figure*}
    \centering
    \includegraphics[width=\linewidth]{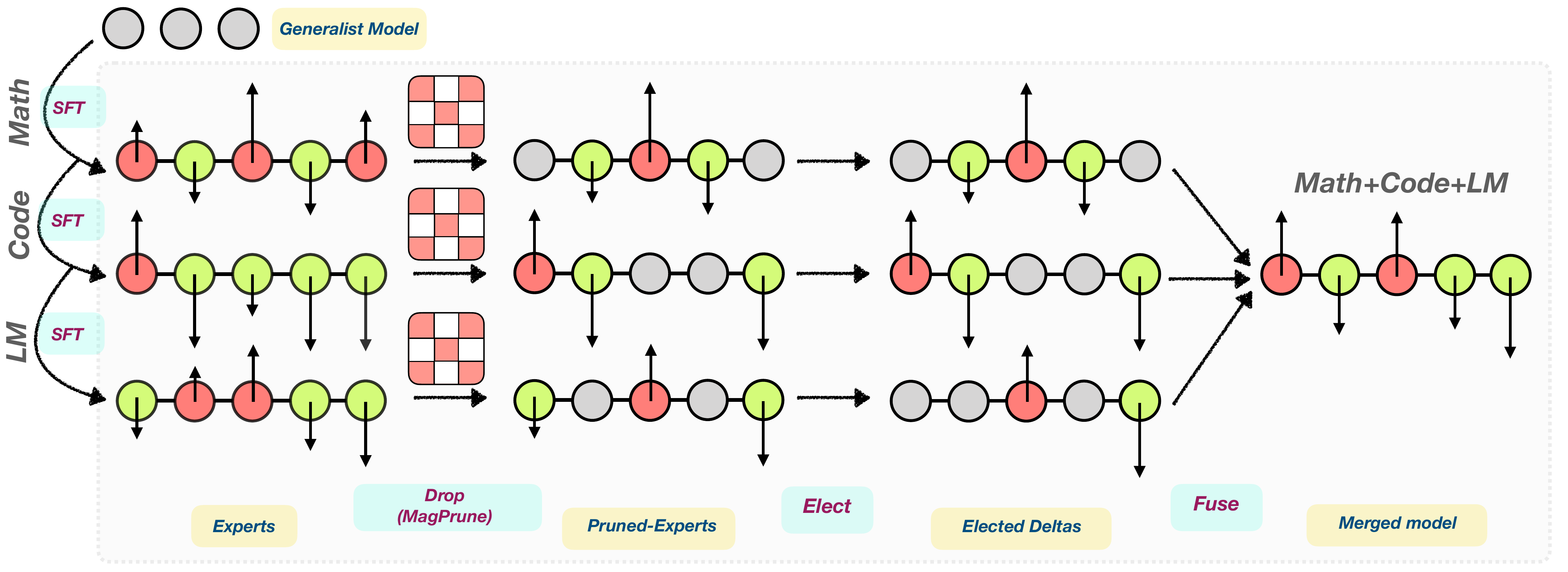}
    \caption{Methodology: Three Steps involved in \approach{}. First step performs magnitude-based sampling of delta parameters (\magp{}), second step elects the parameters that will undergo merging operation, and the final step (Fuse) performs merging.}
    \label{fig:method}
\end{figure*}

On three different homologous (expert) models considered for merging (LM, Math, Code) and their corresponding benchmark datasets (AlpacaEval, GSM8K, MBPP), \approach{} outperforms baseline methods (\dare{}, \ties{}, \ta) in three out of four merges. It achieves an average improvement of 2.4 points over baseline methods employing delta parameter pruning (3.6 points over \ties{}, and 1.2 points over \dare{}), and 11.1 points over the method without pruning (\ta).

Crucial to \approach{} merging, we demonstrate the importance of \magp{}. We observe that \approach{} outperforms the baselines (\random{} in \dare{}, \topk{} in \ties{}, and \nodrop{} in \ta{}) in 5 out of 9 domain-specific performance metrics for merged models, as well as in 2 out of 4 aggregated benchmark scores (Avg.), establishing it as the most effective pruning approach.

We also demonstrate the importance of scaling the unpruned delta parameters. Scaling in \magp{} improves the performance of \approach{} by 7.6 points on the Math+Code model. When the impact of scaling is studied on individual model performance, we observe an average improvement of 28.4 and 28.7 points on \random{} (in \magp{}) and \magp{} (in \approach{}), respectively.

\section{Methodology}
We denote the expert (homologous) models by $\mathcal{M}_t$ with parameters $\bm{\theta}^t$ where $t$ signifies their task expertise. $\mathcal{M}_t$ is obtained from task-specific supervised fine-tuning (SFT) of a base model $\mathcal{M}$ which is common across experts. Let $\bm{\theta}$ denote its parameters. We define delta parameters as the difference between expert and base model parameters, $\bm{\delta}^t \coloneq \bm{\theta}^t - \bm{\theta}$.

\subsection{\approach{}}
The model merging task is to combine multiple experts so that without an additional training phase, the obtained model inherits the qualities of all the experts. \approach{} consists of three steps, 1) Drop: a delta parameter dropping method, that includes a novel magnitude-based pruning approach \textbf{\magp{}}, 2) Elect: Elect the delta parameters that will take part in merging, and 3) Fuse: Perform element-wise addition of delta parameters. The merging process is depicted by \Cref{fig:method}.

\paragraph{Step-1: Drop.} In this step, we set a proportion of delta parameters to zero thus nullifying their role in the expert model \footnote{Dropping a parameter $\delta_i^t$ would mean the parameter value $\theta_i^t$ equals base value $\theta_i$.}. Drop is a crucial step of \approach{} that employs pruning, aimed at lessening the interference between the expert models while preserving the task-specific performance. In this work, we propose a new approach \magp{} that performs delta parameter pruning taking into account their magnitude. For each node in the neural architecture, we map its delta parameters $\{\delta_1, \ldots,\delta_n\}$ to drop probabilities $P_d = \{p_1, \ldots, p_n\}$ in the inverse order of their magnitudes as depicted follows:

\begin{figure}[H]
    \centering
    \includegraphics[width=0.9\linewidth]{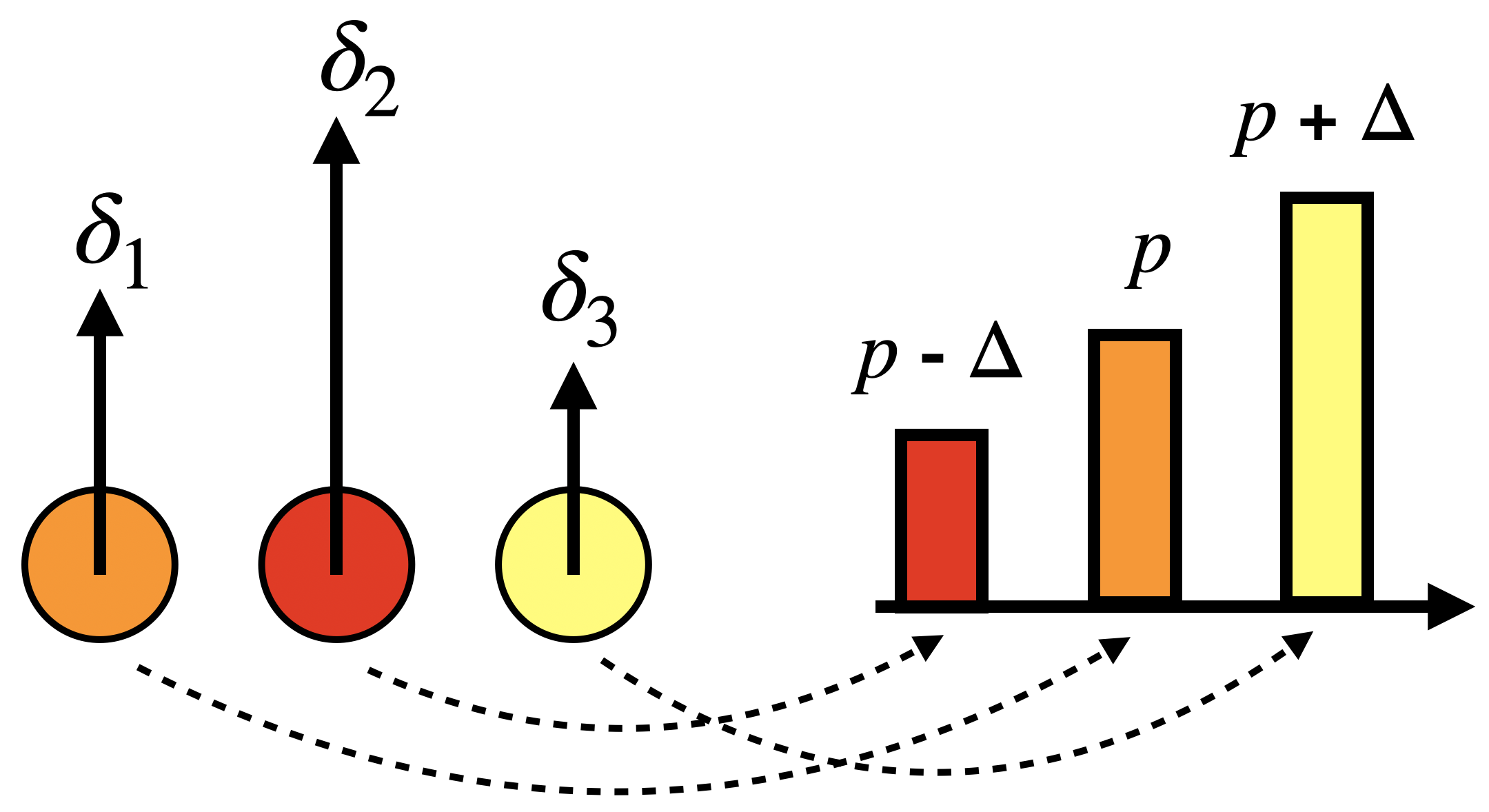}
    \caption{Mapping weights to inversely proportional to drop probabilities.}
    \label{fig:drop}
\end{figure}

where $\Delta$ is a hyperparameter such that $p_{i+1}-p_{i}=\Delta$, we further discuss the choice of hyperparameters in \Cref{step1}.

(\textit{Scaling})---To compensate for the effect of pruning on the model's embedding, we rescale the remaining delta parameters by $1/(1-p_i)$ where $p_i$ is the probability of drop of weight $\delta_i$. We relegate the theoretical analysis of magnitude-based stochastic dropping and scaling to \Cref{step1}. We provide empirical evidence proving the importance of scaling in \Cref{tab:step-1-scaling-expert}.

\paragraph{Step-2: Elect.} Elect refers to selecting delta weights that will undergo the merging operation to minimize further interference caused by merging. Unlike stochastic Drop, Elect reduces the interference by minimizing directional discrepancy in delta parameters. First, we identify the dominating direction by noticing the sign of the sum of all the delta parameters $S = \text{sgn} (\sum_{t=1}^{T} \delta_k^{t})$ at position $k$. Thus, the elected delta parameters for position $k$ are ones carrying the same sign as $S$:
\begin{equation*}
    C = \{t \in [T]\;|\;\text{sgn}(\delta^t_k)=S\},
\end{equation*}
where $C$ is the index of such delta parameters.

\paragraph{Step-3: Fuse.} We obtain the merged delta parameter at position $k$ by computing the average of elected delta parameters, i.e.,
\begin{align*}
    \delta_{k}^{avg} = \sum_{i\in C} \delta_{k}^i
\end{align*}
However, similar to Step 1, to compensate for the drop in model parameters, we rescale the delta values to obtain the final merged model $\bm{\theta}_{m}$:
\begin{equation}
    \bm{\theta}_m = \bm{\theta} +\lambda*\bm{\delta}^{avg},
\end{equation}
where $\lambda$ is a scalar value. We provide an analysis on $\lambda$ in \Cref{fig:LambdaSweep}.

\subsection{\magp{}: Stochastic Magnitude-based Pruning} \label{step1}
As a critical step of merging, we detail the sampling step. For a set of delta parameters in an expert model $\delta_1, \delta_2, \ldots\delta_n$, first we rank them followed by assigning them to corresponding drop probabilities~\footnote{We perform the ranking of the delta parameters row-wise fashion, as it performed better than layer-wise ranking as seen in \Cref{fig:LayervsRow}.}:
\begin{align*}
    \{r_1, r_2, \ldots, r_n\} &= \text{rank}(\{\delta_1, \delta_2, \ldots\delta_n\})\\
    \Delta_i &= \frac{\epsilon}{n}*r_i\\
    p_i &= p_{min} + \Delta_i
\end{align*}

where $r_i \in \{0, 1,\ldots,n-1\}$ and $p_i$ is the rank, and drop probability of $i_{th}$ delta; $p_{min}$ is the minimum probability of dropping assigned to maximum magnitude delta parameter such that
\[
p_{min} = p - \frac{\epsilon}{2}
\]

where $p$ is the average drop probability. Next, we perform the sampling step:
\begin{align*}
    m_i &\sim \text{Bernoulli}(p_i),\\
    \tilde{\delta}_i &= (1-m_i) \odot \delta_i,\\
    \hat{\delta}_i &= \frac{\tilde{\delta}_i}{1-p_i}.
\end{align*}
Here $m_i$ value 1 would denote $\delta_i$ is dropped, $1-p_i$ denotes rescaling which is effective to all the undropped delta parameters. Next, we provide the theoretical justification of the scaling factor.

\paragraph{\dare{} and \ties{} as configuration of \approach{}.} Notably, in \magp{}, while we define $P_s$ with step probabilities $\Delta{}$, one can freely choose any set of valid probabilities $p_i's$ to map $\delta's$, to explore more possibilities of dropping. To obtain pruning used in \dare{}, one can choose all $p_i$'s to be the same. To obtain pruning used in \ties{} \footnote{Keep top-K magnitudes.}, one can assign 1 to K elements in $P_d$ and 0 to others.

\begin{algorithm}
\caption{Drop-Elect-Merge with \approach{}.}
\label{alg:ties-merging}
\renewcommand{\algorithmicrequire}{\textbf{Input:}}
\renewcommand{\algorithmicensure}{\textbf{Output:}}
\begin{algorithmic}
\Require Fine-tuned model parameters $\{\bm{\theta}^t\}_{t=1}^T$, Pre-trained model parameters $\bm{\theta}$, $p$, $\lambda$ and $\epsilon$.
\Ensure Merged Model Parameters $\bm{\theta}_{m}$
\ForAll {$t$ in $1, \ldots, T$}
    \State \(\triangleright\) \textbf{Step 0} Get Delta parameters
    \State $\bm{\delta}^t = \bm{\theta}^t - \bm{\theta}$\\
    \State \(\triangleright\) \textbf{Step 1:} Drop 
    \State $\bm{\hat{\delta}}^t \gets \text{\magp{}}(\bm{\delta}^t, p, \epsilon)$
    \State $\bm{\gamma}^t \gets \text{sign}(\bm{\hat{\delta}}^t)$
    \State $\bm{\mu}^t \gets |\bm{\hat{\delta}}^t|$
\EndFor\\ \\
\(\triangleright\) \textbf{Step 2:} Elect 
\State $\bm{\gamma}^{m} = \text{sign}(\sum_{t=1}^T \bm{\hat{\delta}}^t)$\\ \\
\(\triangleright\) \textbf{Step 3:} Merge 
\ForAll {$k$ in $1, \ldots, d$}
    \State $C_k = \{t \in [T] | \gamma^t_k = \gamma^{m}_k\}$
    \State $\delta^{m}_k = \frac{1}{|C_k|} \sum_{t \in C_k} \hat{\delta}^t_k$
\EndFor\\ \\
\(\triangleright\) \textbf{Step 4:} Obtain merged checkpoint
\State $\bm{\theta}_{m} \gets \bm{\theta} + \lambda * \bm{\delta}^{m}$
\State \Return $\bm{\theta}_m$
\end{algorithmic}
\end{algorithm}

\subsection{Theoretical Analysis} 
We analyze the impact of Drop step on the linear transformations. Let $h$ denote a node in the network with weights $w_1, \ldots, w_n$ and corresponding delta parameters $\delta_1, \ldots, \delta_n$, the input embedding vector be $\{x1, \ldots, x_n\}$. We can define expected output embedding as:
\begin{align*}
    \mathbb{E}[h] &= \mathbb{E}\left[\sum_{i=1}^{n} \left(w_{i} + \delta{i}\right) x_i\right]\\
    &= \sum_{i=1}^{n} x_i \mathbb{E}[w_{i}] + \sum_{i=1}^{n} x_i \mathbb{E}[\delta {i}]\\
    &= \sum_{i=1}^{n} x_i w_{i} + \sum_{i=1}^{n} x_i \delta {i} = h + \Delta h\\
\end{align*}
In Step 1 of \approach{}, we drop the delta parameters with a probability $p_i$ and rescale others by $\gamma$, the expected output of the node can be identified as
\begin{align*}
    \mathbb{E}[\hat{h}] &= \mathbb{E}\left[\sum_{i=1}^{n} \left(w_{i} + \hat{\delta}{i}\right) x_i\right]\\
    &= \sum_{i=1}^{n} x_i \mathbb{E}[w_{i}] + \sum_{i=1}^{n} x_i \mathbb{E}[\hat{\delta} {i}]\\
     &= \sum_{i=1}^{n} x_i w_{i} + \sum_{i=1}^{n} x_i (1-p_i)*\gamma*\delta {i}\\
     &= h + \sum_{i=1}^{n} x_i*(1-p_i)*\gamma*\delta {i}
\end{align*}
If we put $\gamma = 1/(1-p_i)$, we get,
\begin{align*}
    \mathbb{E}[\hat{h}] &= h + \sum_{i=1}^{n} x_i*\delta {i} \\
    \mathbb{E}[\hat{h}] &= h + \Delta h
\end{align*}

\begin{figure*}
    \centering
    \includegraphics[width=1\linewidth]{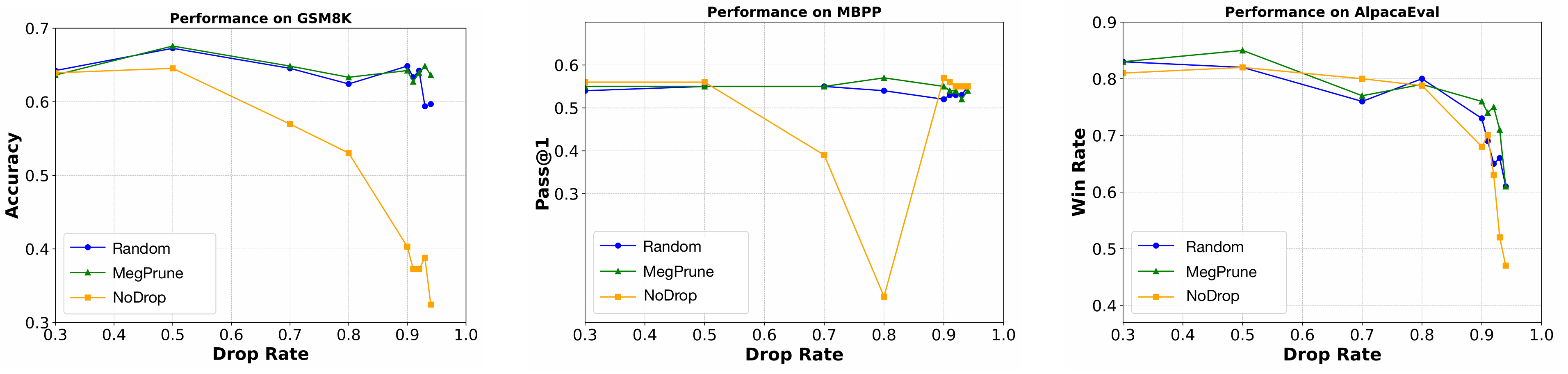}
    \caption{Performance vs Drop Rate comparison of \approach{} (magnitude-based random drop) against baselines \dare{} (random drop) and \ties{} (magnitude-based deterministic drop).}
    \label{fig:compare_pruning}
\end{figure*}

Thus, similar to \dare{}, \approach{} can preserve model performance by approximating the original embeddings, yet being a more generic (superset) pruning approach. \Cref{alg:ties-merging} outlines the steps of \approach{} with the Drop-Elect-Merge approach.

\section{Experimental Setup}
\paragraph{Expert Models.} We experiment with the Wizard model-based experts. WizardLM (\textbf{LM-expert}) \cite{xu2024wizardlm} for instruction following, WizardMath \cite{luo2023wizardmath} for math (\textbf{Math-expert}), and WizardCoder-Python for code (\textbf{Code-expert}). Specific to model merging, a primary requirement is to have experts obtained from the same backbone model. Since WizardCoder-Python \cite{luo2024wizardcoder} uses a different pre-trained backbone (CodeLlama), we use llama-2-13b-code-alpaca \footnote{https://huggingface.co/layoric/llama-2-13b-code-alpaca} as code expert which shares a common backbone with WizardLM and WizardMath i.e. LLaMA-2.

\paragraph{Baselines.} We compare \approach{} with two state-of-the-art merging approaches: \dare{} and \ties{}. {\color{darkgray} \textbf{For valid comparisons, we keep Step-2 (Elect) and Step-3 (Merge) consistent across all baselines and compare them in Step-1, which involves interference resolution in delta parameters.}}

\begin{itemize}
    \item  \textbf{\ta{}} \cite{ilharco2023editing}---Task Arithmetic (\ta{}) is a simple delta weight merging without Step-1 and 2. However, to provide a valid comparison, we employ Step-2 i.e. Elect. We refer to the absence of dropping by \textbf{\nodrop{}}.

    \item  \textbf{\dare{}} \cite{dare}--- \dare{} randomly drops the delta parameter at a uniform drop rate, $p$, and rescales the remaining values by $1/(1-p)$, referred to as \textbf{\random{}}.
    
    \item  \textbf{\ties{}} \cite{ties}--- \ties{} retains only the top-K parameters with the highest magnitudes in each layer, referred to as \textbf{\topk{}}.
    
    \item \textbf{\approach{}} (Ours)--- In Step-1, \approach{} employs \textbf{\magp{}} that first ranks delta magnitudes (row-wise), followed by assigning them to a predefined set of probabilities.
\end{itemize}

(\textit{Hyperparameter Search})---To achieve the best performance with all the baseline approaches, we conducted a hyperparameter search to determine the suitable values of the drop rate $p$ and rescaling factor, $\lambda$, wherever applicable. We performed merging by varying drop rates \textit{p} in [0.1, 0.3, 0.5, 0.7, 0.9]  and selected the optimal \textit{p} for each approach to compare the final performance. We evaluated the merged models' performance by evaluating them on all the tasks of its base models. We set the temperature parameter to 0.0 for greedy decoding and set the maximum number of generated tokens to 512. We used the mergekit library \cite{mergekit} to perform the pruning and merging operations on the model parameters.

(\textit{Casting \approach{} as \dare{} and \ties{}})---Essentially, \approach{} encompasses NoDrop, \dare{}, \ties{}, and rank-proportional probability distribution. \textbf{\approach{}$_{p=0}$} is essentially no dropping of parameters given the probability of dropping each parameter sampled is 0; \textbf{\approach{}$_{\epsilon=0}$} boils down to \dare{} with all the parameters having the same probability $p_a$ of getting sampled (mentioned in \Cref{step1}). \textbf{\approach$_{p=0/1}$} refers to mapping the ranked delta magnitudes to probabilities where $k$ probabilities are 1s and rest are $(n-k)$ are 0s. \textbf{\approach$_{p \propto r}$} is magnitude proportional sampling of delta parameters.

\paragraph{Evaluation Metrics.} We used AlpacaEval \cite{alpaca_eval} to evaluate the performance of models on the instruction-following task, GSM8K to evaluate a model's performance on mathematical reasoning \cite{cobbe2021gsm8k}, and MBPP \cite{mbpp} for evaluation of code generation. For AlpacaEval, we compare the model-generated answer against the reference answer using GPT4 as an evaluator to determine the win rate of the model against the reference solution. For MBPP, we use the Pass@1 score to measure the performance of the model's code generation capabilities. Lastly, for GSM8K, we use GPT4-as-a-judge to evaluate the mathematical correctness i.e. the final answer as well as the intermediate steps (reasoning).

\paragraph{Experiments (w/o Elect).} We also cast \ta{}, \dare{}, \ties{}, and \approach{} in a setup where we remove the participation of Step-2 i.e. Elect. This setting is the standard setting used in \ta{} \cite{ilharco2023editing} and \dare{}\cite{dare}. 

\paragraph{Experiments (only Drop).} We also compare different pruning methods, \random{}, \topk{}, and \magp{}, directly on expert models.

\begin{table}[h]
\centering
\resizebox{0.5\textwidth}{!}{%
\begin{tabular}{lllcccc}
\hline
\textbf{Merging} &
  \multicolumn{1}{c}{\textbf{Models}} &
  \multicolumn{1}{c}{\textbf{Method}} &
  \textbf{AlpacaEval} &
  \textbf{GSM8K} &
  \textbf{MBPP} &
  \textbf{Avg.} \\ \hline
\multirow{3}{*}{Single} &
  LM &
  \multicolumn{1}{c}{-} &
  80.8 &
  45.9 &
  \multicolumn{1}{c|}{31.4} &
  / \\
 &
  Math &
  \multicolumn{1}{c}{-} &
  / &
  63.5 &
  \multicolumn{1}{c|}{/} &
  / \\
 &
  Code &
  \multicolumn{1}{c}{-} &
  / &
  / &
  \multicolumn{1}{c|}{27.0} &
  / \\ \hline
\multirow{28}{*}{Merge} &
  \multirow{7}{*}{\begin{tabular}[c]{@{}l@{}}LM + \\ Math\end{tabular}} &
  \begin{tabular}[c]{@{}l@{}}\ta\\ (\approach$_{p=0}$)\end{tabular} &
  79.1 &
  60.6 &
  \multicolumn{1}{c|}{/} &
  69.9 \\ \cline{3-7} 
 &
   &
  \begin{tabular}[c]{@{}l@{}}\dare{}\\ (\approach$_{\epsilon=0}$)\end{tabular} &
  81.3 &
  63.3 &
  \multicolumn{1}{c|}{/} &
  72.3 \\ \cline{3-7} 
 &
   &
  \begin{tabular}[c]{@{}l@{}}\ties{}\\ (\approach$_{p=0/1}$)\end{tabular} &
  77.6 &
  \textbf{66.8} &
  \multicolumn{1}{c|}{/} &
  72.2 \\ \cline{3-7} 
 &
   &
  \renewcommand{\arraystretch}{2}\begin{tabular}[c]{@{}l@{}}\approach$_{p \propto r}$\end{tabular} &
  \textbf{81.8} &
  64.1 &
  \multicolumn{1}{c|}{/} &
  \textbf{72.9} \\ \cline{2-7} 
 &
  \multirow{7}{*}{\begin{tabular}[c]{@{}l@{}}LM + \\ Code\end{tabular}} &
  \begin{tabular}[c]{@{}l@{}}\ta\\ (\approach$_{p=0}$)\end{tabular} &
  73.3 &
  / &
  \multicolumn{1}{c|}{0} &
  36.7 \\ \cline{3-7} 
 &
   &
  \begin{tabular}[c]{@{}l@{}}\dare{}\\ (\approach$_{\epsilon=0}$)\end{tabular} &
  79.4 &
  / &
  \multicolumn{1}{c|}{32.8} &
  56.1 \\ \cline{3-7} 
 &
   &
  \begin{tabular}[c]{@{}l@{}}\ties{}\\ (\approach$_{p=0/1}$)\end{tabular} &
  77.4 &
  / &
  \multicolumn{1}{c|}{9.6} &
  43.5 \\ \cline{3-7} 
 &
   &
  \renewcommand{\arraystretch}{2}\begin{tabular}[c]{@{}l@{}}\approach$_{p \propto r}$\end{tabular} &
  \textbf{79.7} &
  / &
  \multicolumn{1}{c|}{\textbf{35.2}} &
  \textbf{57.5} \\ \cline{2-7} 
 &
  \multirow{7}{*}{\begin{tabular}[c]{@{}l@{}}Math + \\ Code\end{tabular}} &
  \begin{tabular}[c]{@{}l@{}}\ta\\ (\approach$_{p=0}$)\end{tabular} &
  / &
  63.2 &
  \multicolumn{1}{c|}{15.6} &
  39.4 \\ \cline{3-7} 
 &
   &
  \begin{tabular}[c]{@{}l@{}}\dare{}\\ (\approach$_{\epsilon=0}$)\end{tabular} &
  / &
  61.9 &
  \multicolumn{1}{c|}{20.0} &
  41.0 \\ \cline{3-7} 
 &
   &
  \begin{tabular}[c]{@{}l@{}}\ties{}\\ (\approach$_{p=0/1}$)\end{tabular} &
  / &
  \textbf{65.9} &
  \multicolumn{1}{c|}{\textbf{20.8}} &
  \textbf{43.4} \\ \cline{3-7} 
 &
   &
  \renewcommand{\arraystretch}{2}\begin{tabular}[c]{@{}l@{}}\approach$_{p \propto r}$\end{tabular} &
  / &
  64.9 &
  \multicolumn{1}{c|}{20.0} &
  42.4 \\ \cline{2-7} 
 &
  \multirow{7}{*}{\begin{tabular}[c]{@{}l@{}}LM + \\ Math + \\ Code\end{tabular}} &
  \begin{tabular}[c]{@{}l@{}}\ta\\ (\approach$_{p=0}$)\end{tabular} &
  72.1 &
  48.3 &
  \multicolumn{1}{c|}{0} &
  40.1 \\ \cline{3-7} 
 &
   &
  \begin{tabular}[c]{@{}l@{}}\dare{}\\ (\approach$_{\epsilon=0}$)\end{tabular} &
  77.5 &
  62.7 &
  \multicolumn{1}{c|}{29.4} &
  56.5 \\ \cline{3-7} 
 &
   &
  \begin{tabular}[c]{@{}l@{}}\ties{}\\ (\approach$_{p=0/1}$)\end{tabular} &
  77.5 &
  \textbf{66.9} &
  \multicolumn{1}{c|}{27.2} &
  57.2 \\ \cline{3-7} 
 &
   &
  \renewcommand{\arraystretch}{2}\begin{tabular}[c]{@{}l@{}}\approach$_{p \propto r}$\end{tabular} &
  \textbf{80.4} &
  61.8 &
  \multicolumn{1}{c|}{\textbf{31.4}} &
  \textbf{57.9} \\ \hline
\end{tabular}
}
\caption{\approach{} against baseline approaches in Drop-Elect-Fuse Setup.}
\label{tab:three_step_results}
\end{table}

\begin{table}[h]
\centering
\resizebox{\linewidth}{!}{%
\begin{tabular}{@{}lccc@{}}
\toprule
\textbf{Method}          & \textbf{GSM8K} & \textbf{MBPP} & \textbf{Avg.} \\ \midrule
\approach (w/o scaling) & 42.2        & \textbf{25.6}         & 33.9        \\
\approach (w/ scaling)              & \textbf{63.8} $({\uparrow 21.6})$   & 19.2 $({\downarrow -6.4})$ & \textbf{41.5} $({\uparrow 7.6})$ \\ \bottomrule
\end{tabular}%
}
\caption{Effect of scaling on merging (at $\lambda=1$).}
\label{tab:step-1-scaling}
\end{table}

\section{Results}

\paragraph{Merging.} Shown in \Cref{tab:three_step_results}, \approach$_{p \propto r}$ outperforms the baselines \dare{}, \ties{}, and \ta{} in three out of four merge settings: LM+Math ($\uparrow$ 0.6 points), LM+Code ($\uparrow$ 1.4 points), and LM+Math+Code ($\uparrow$ 0.7 points). It is on the GSM8K (Math) task where merging models using \approach$_{p \propto r}$, as well as \dare{}, do not tend to work as well as \ties{}. We posit this is due to the randomness in the delta parameter dropping that causes performance drops. This is likely because the coding expert does not have enough task-prominent parameters to sustain the inference from other experts (LM and Code), despite considerable performance retention even after pruning 90\% of the delta parameters.

Using \approach$_{p \propto r}$, merging the LM-expert with other experts (LM+<other-experts>) is shown to maintain performance on the AlpacaEval task, scoring within approximately $\pm$1 point of the LM-expert's score on the task (80.8), with an average drop of $\downarrow$0.2 points. The average drop in performance using \ties{} and \dare{} merging is observed to be higher, i.e., $\downarrow$3.3 points and $\downarrow$1.4 points, respectively. Without employing pruning Step-1, i.e., \ta{} setting observes the highest performance degradation of $\downarrow$5.9 across baselines. Merging the Math-expert with others (Math+<other-experts>) generally improves performance on GSM8K (\approach{} shows an average increase of $\uparrow$0.1 points), whereas the three-expert setting is observed to face interference, diminishing the scores post-merging. Merging the Code-expert with others (Code+<other-experts>) shows an increase in performance in two out of three merges, with an average improvement of $\uparrow$2.1 points on MBPP Python problems compared to the Code-only expert. Whereas \dare{} observes a smaller increase of $\uparrow$0.4 points and \ties{} faces a decrease in score by $\downarrow$-7.8 points. \ta{}'s scored zero on the MBPP task whenever Code-expert is merged with LM-expert.

A similar case occurs with \ties{}, which shows a significant drop of 17.4 on the MBPP coding task in the LM+Code merge setting. Upon closer examination, we observe a greater influence of the LM-expert on the model, making the outputs more chat-like rather than structured code-like. We believe scaling (not present in \ties{}) plays an important role here; not scaling (up) the parameters after dropping can cause dominating effects, where one expert overshadows the qualities of the other.

\begin{figure}
    \centering
    \includegraphics[width=1\linewidth]{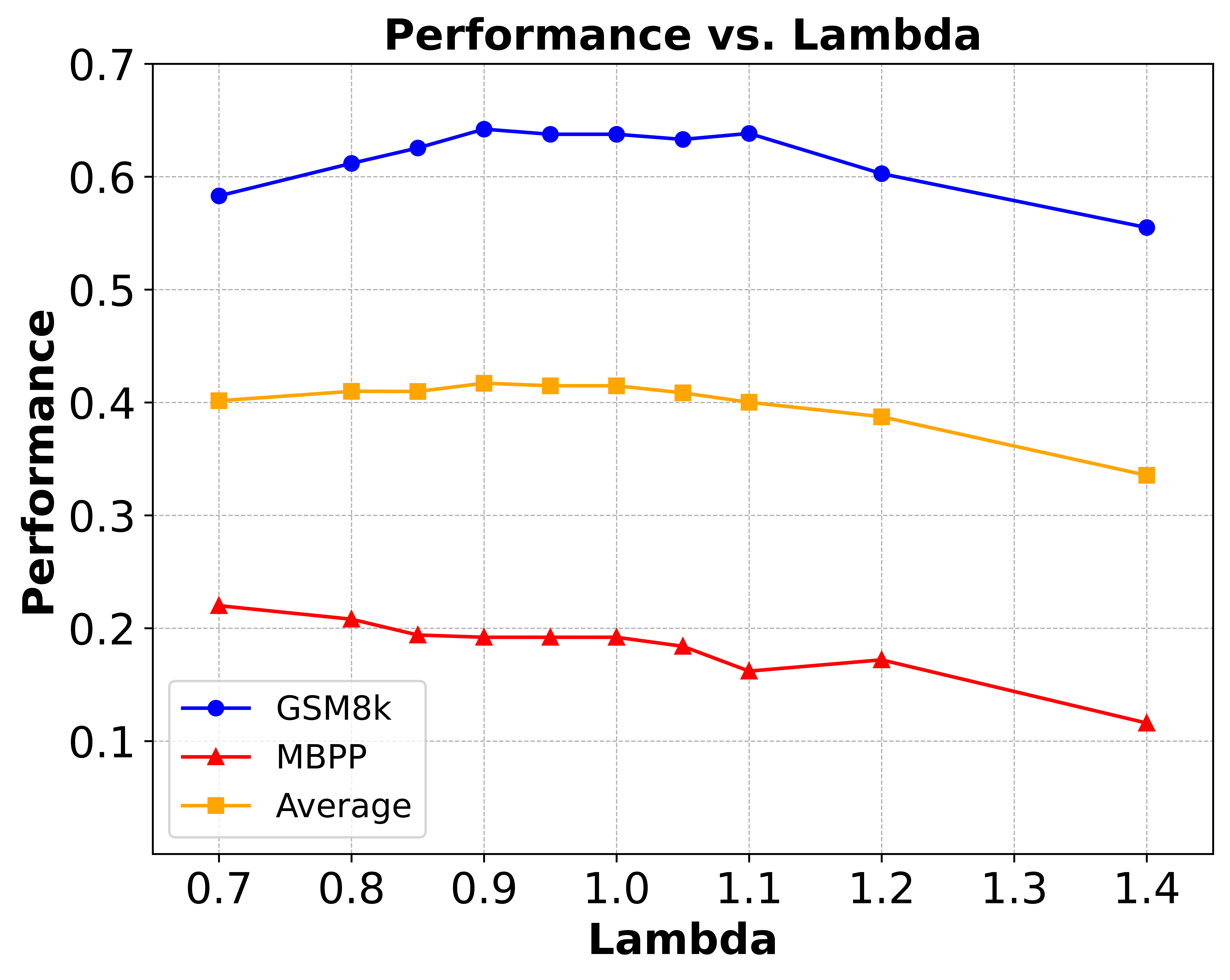}
    \caption{Performance vs lambda for the math+code merge combination}
    \label{fig:LambdaSweep}
\end{figure}

\paragraph{Effect of scaling.} To understand the importance of $1/(1-p)$ scaling in Step-1, we prune expert models\footnote{Unless mentioned, \approach{} refers to \approach$_{p \propto r}$.} using different drop methods and understand the scaling effect on individual models' performance as well as merged models. \Cref{tab:step-1-scaling} shows the effect of scaling after Step-1 on the Math+Code model, we observe the scaling raises the score on the coding task (GSM8K) with an increase in score by $\uparrow$21.6 points while a $\uparrow$6.4 point reduction was observed on MBPP math task. Overall, scaling improved the average score of the merged experts by $\uparrow$7.6 points. A similar observation is made when we analyse the importance of scaling on pruned expert models with at drop rate $p$ \Cref{tab:step-1-scaling-expert}. We observe an average of improvement by $\uparrow$28.4 and $\uparrow$28.7 points on \dare{} and \approach{}, respectively.

\begin{table}[]
\centering
\resizebox{\linewidth}{!}{%
\begin{tabular}{@{}ccccc@{}}
\toprule
\begin{tabular}[c]{@{}c@{}}Sparsity\\ ($p$)\end{tabular} &
  \begin{tabular}[c]{@{}c@{}}\dare{} \\ (w/o scaling)\end{tabular} &
  \begin{tabular}[c]{@{}c@{}}\dare{}\\ (w/ scaling)\end{tabular} &
  \begin{tabular}[c]{@{}c@{}}\approach \\ (w/o scaling)\end{tabular} &
  \begin{tabular}[c]{@{}c@{}}\approach\\ (w/ scaling)\end{tabular} \\ \midrule
0.3 & 39.7 & \multicolumn{1}{c|}{\textbf{48.5} ${(\uparrow 8.8)}$}  & 42.1 & \textbf{47.3} ${(\uparrow 5.2)}$  \\
0.5 & 33.7 & \multicolumn{1}{c|}{\textbf{47.3} ${(\uparrow 13.6)}$} & 34.9 & \textbf{48.5} ${(\uparrow 13.6)}$ \\
0.7 & 13.0 & \multicolumn{1}{c|}{\textbf{43.3} ${(\uparrow 30.3)}$} & 19.7 & \textbf{48.8} ${(\uparrow 29.1)}$ \\
0.8 & 2.4  & \multicolumn{1}{c|}{\textbf{45.5} ${(\uparrow 43.1)}$} & 2.4  & \textbf{49.1} ${(\uparrow 46.7)}$ \\
0.9 & 0    & \multicolumn{1}{c|}{\textbf{46.4} ${(\uparrow 46.4)}$} & 0.3  & \textbf{49.1} ${(\uparrow 48.8)}$ \\ \bottomrule
\end{tabular}%
}
\caption{Effect of scaling on expert's performance.}
\label{tab:step-1-scaling-expert}
\end{table}

\paragraph{Results on w/o Elect.} In \Cref{tab:TA_results}, we focus on comparing the effect of different pruning approaches on merging, thus skipping Step 2, i.e., Elect. On average scores, we observe that \approach{} outperforms the baselines in 5 out of 9 merged model-benchmark task settings, as well as 2 out of 4 times on aggregated benchmark scores (Avg.). We also notice that the performance drop due to omitting Step 2 (Elect) is the least for \approach{} compared to the other baselines. This shows \magp{} drop in \approach{} intrinsically facilitates electing the parameters.

\paragraph{How effective is \magp{}?} We study the effect of \magp{} on individual expert models (\Cref{fig:compare_pruning}). We find \magp{} consistently outperforms \topk{} on math task GSM8K with a wider performance gap at higher pruning ratios. A similar observation is made on the AlpacaEval task where except for the pruning ratio 0.7, \magp{} outperforms \topk{}. \magp{} outperforms \random{} on AlpacaEval consistently except for $p=0.8$. On math task MBPP, we observe an inconsistent trend, making it hard to identify a comprehensive winner. One thing to note from GSM8K and AlpacaEval results is that \magp{} becomes more effective at higher ratios $p\geq0.8$, preserving more expert-specific capabilities while identifying more crucial parameters to solve the task. 

\begin{figure}[h]
    \centering
    \includegraphics[width=1\linewidth]{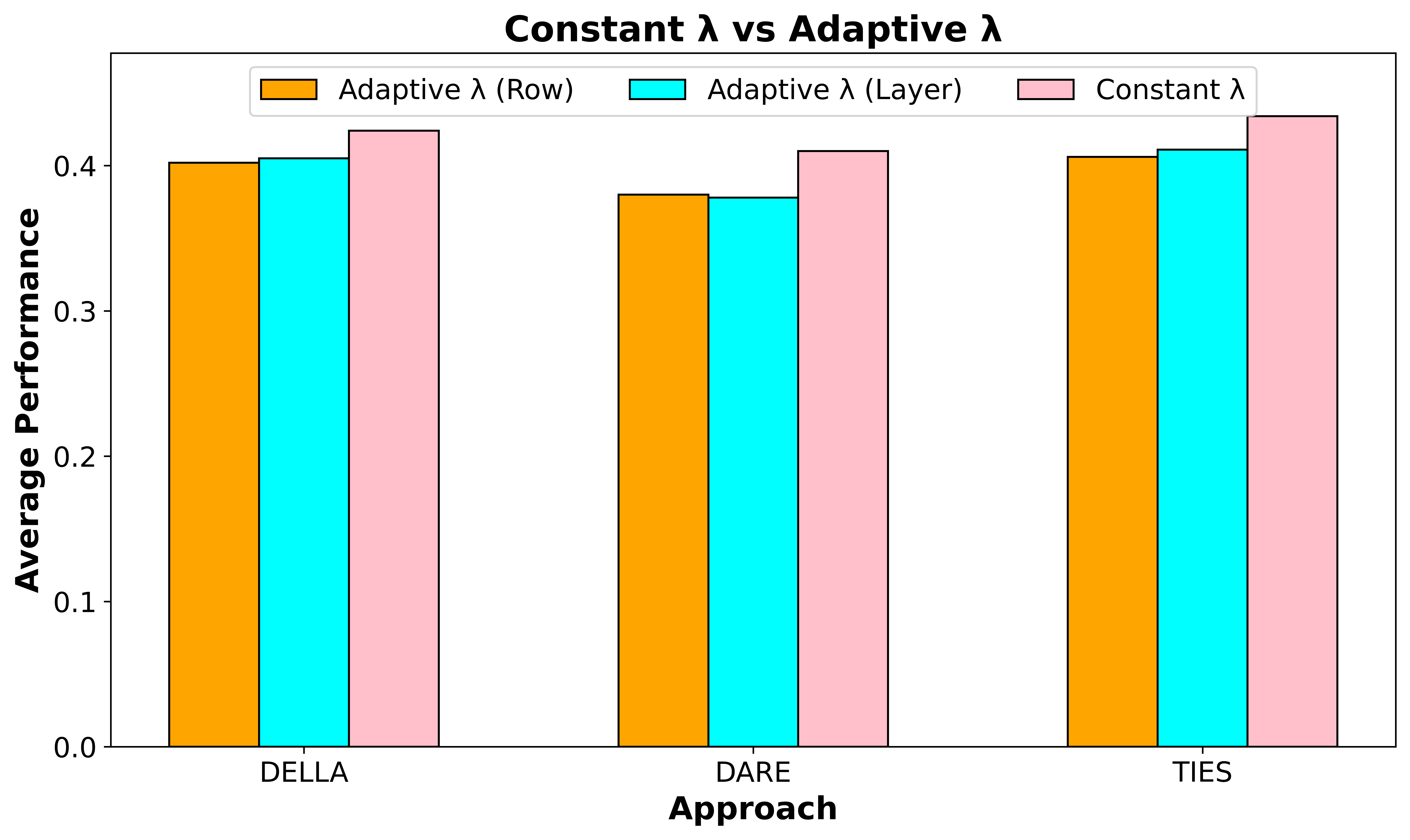}
    \caption{\footnotesize{Adaptive Scaling vs Constant lambda Scaling \approach{}}}
    \label{fig:lambdaScaling comparison}
\end{figure}

\begin{table}[h]
\centering
\resizebox{0.48\textwidth}{!}{%
\begin{tabular}{lllcccc}
\hline
\textbf{Merging} &
  \multicolumn{1}{c}{\textbf{Models}} &
  \multicolumn{1}{c}{\textbf{Method}} &
  \textbf{AlpacaEval} &
  \textbf{GSM8K} &
  \textbf{MBPP} &
  \textbf{Avg.} \\ \hline
\multirow{3}{*}{Single} &
  LM &
  \multicolumn{1}{c}{-} &
  80.8 &
  45.9 &
  \multicolumn{1}{c|}{31.4} &
  / \\
 &
  Math &
  \multicolumn{1}{c}{-} &
  / &
  63.5 &
  \multicolumn{1}{c|}{/} &
  / \\
 &
  Code &
  \multicolumn{1}{c}{-} &
  / &
  / &
  \multicolumn{1}{c|}{27.0} &
  / \\ \hline
\multirow{28}{*}{Merge} &
  \multirow{7}{*}{\begin{tabular}[c]{@{}l@{}}LM + \\ Math\end{tabular}} &
  \begin{tabular}[c]{@{}l@{}}\ta\\ (\approach$_{p=0}$)\end{tabular} &
  73.7 &
  65.2 &
  \multicolumn{1}{c|}{/} &
  69.5 \\ \cline{3-7} 
 &
   &
  \begin{tabular}[c]{@{}l@{}}\dare{}\\ (\approach$_{\epsilon=0}$)\end{tabular} &
  72.7 &
  64.8 &
  \multicolumn{1}{c|}{/} &
  68.9 \\ \cline{3-7} 
 &
   &
  \begin{tabular}[c]{@{}l@{}}\ties{}\\ (\approach$_{p=0/1}$)\end{tabular} &
  72.1 &
  \textbf{65.7} &
  \multicolumn{1}{c|}{/} &
  68.9 \\ \cline{3-7} 
 &
   &
  \renewcommand{\arraystretch}{2}\begin{tabular}[c]{@{}l@{}}\approach$_{p \propto r}$\end{tabular} &
  \textbf{75.0} &
  65.5 &
  \multicolumn{1}{c|}{/} &
  \textbf{70.2} \\ \cline{2-7} 
 &
  \multirow{7}{*}{\begin{tabular}[c]{@{}l@{}}LM + \\ Code\end{tabular}} &
  \begin{tabular}[c]{@{}l@{}}\ta\\ (\approach$_{p=0}$)\end{tabular} &
  67.2 &
  / &
  \multicolumn{1}{c|}{33.2} &
  50.2 \\ \cline{3-7} 
 &
   &
  \begin{tabular}[c]{@{}l@{}}\dare{}\\ (\approach$_{\epsilon=0}$)\end{tabular} &
  \textbf{67.6} &
  / &
  \multicolumn{1}{c|}{33.0} &
  \textbf{50.3} \\ \cline{3-7} 
 &
   &
  \begin{tabular}[c]{@{}l@{}}\ties{}\\ (\approach$_{p=0/1}$)\end{tabular} &
  64.9 &
  / &
  \multicolumn{1}{c|}{34.6} &
  49.8 \\ \cline{3-7} 
 &
   &
  \renewcommand{\arraystretch}{2}\begin{tabular}[c]{@{}l@{}}\approach$_{p \propto r}$\end{tabular} &
  65.7 &
  / &
  \multicolumn{1}{c|}{\textbf{34.8}} &
  50.2 \\ \cline{2-7} 
 &
  \multirow{7}{*}{\begin{tabular}[c]{@{}l@{}}Math + \\ Code\end{tabular}} &
  \begin{tabular}[c]{@{}l@{}}\ta\\ (\approach$_{p=0}$)\end{tabular} &
  / &
  51.2 &
  \multicolumn{1}{c|}{7.8} &
  \textbf{29.6} \\ \cline{3-7} 
 &
   &
  \begin{tabular}[c]{@{}l@{}}\dare{}\\ (\approach$_{\epsilon=0}$)\end{tabular} &
  / &
  48.8 &
  \multicolumn{1}{c|}{7.0} &
  27.9 \\ \cline{3-7} 
 &
   &
  \begin{tabular}[c]{@{}l@{}}\ties{}\\ (\approach$_{p=0/1}$)\end{tabular} &
  / &
  \textbf{52.9} &
  \multicolumn{1}{c|}{5.6} &
  29.3 \\ \cline{3-7} 
 &
   &
  \renewcommand{\arraystretch}{2}\begin{tabular}[c]{@{}l@{}}\approach$_{p \propto r}$\end{tabular} &
  / &
  48.7 &
  \multicolumn{1}{c|}{\textbf{8.6}} &
  28.6 \\ \cline{2-7} 
 &
  \multirow{7}{*}{\begin{tabular}[c]{@{}l@{}}LM + \\ Math + \\ Code\end{tabular}} &
  \begin{tabular}[c]{@{}l@{}}\ta\\ (\approach$_{p=0}$)\end{tabular} &
  62.2 &
  55.8 &
  \multicolumn{1}{c|}{0} &
  40.1 \\ \cline{3-7} 
 &
   &
  \begin{tabular}[c]{@{}l@{}}\dare{}\\ (\approach$_{\epsilon=0}$)\end{tabular} &
  63.1 &
  55.0 &
  \multicolumn{1}{c|}{30.2} &
  49.4 \\ \cline{3-7} 
 &
   &
  \begin{tabular}[c]{@{}l@{}}\ties{}\\ (\approach$_{p=0/1}$)\end{tabular} &
  63.6 &
  57.7 &
  \multicolumn{1}{c|}{32.2} &
  51.2 \\ \cline{3-7} 
 &
   &
  \renewcommand{\arraystretch}{2}\begin{tabular}[c]{@{}l@{}}\approach$_{p \propto r}$\end{tabular} &
  \textbf{77.5} &
  \textbf{66.8} &
  \multicolumn{1}{c|}{29.8} &
  \textbf{58.0} \\ \hline
\end{tabular}%
}
\caption{Comparison of \approach{} against baselines in merging w/o Elect setting.}
\label{tab:TA_results}
\end{table}

\paragraph{Adaptive vs Constant $\bm{\lambda}$?}
\Cref{fig:lambdaScaling comparison} shows that applying constant lambda scaling after merging the delta weights (after Fusion step 3) results in greater performance improvements for the merged model compared to adaptively scaling each model's deltas based on the proportion of parameters elected in Step 2 (effective $p$ after Step 2). Using Constant $\lambda$ scaling leads to a notable improvement across all merging approaches: \approach{} ($\uparrow$ 1.5 points), \dare{} ($\uparrow$ 2.9 points), \ties{} ($\uparrow$ 2.3 points). To find the optimal value of the constant $\lambda$, we perform a hyperparameter search and select the best performing $\lambda$ as shown in \Cref{fig:LambdaSweep}.

\begin{figure}
    \centering
    \includegraphics[width=0.8\linewidth]{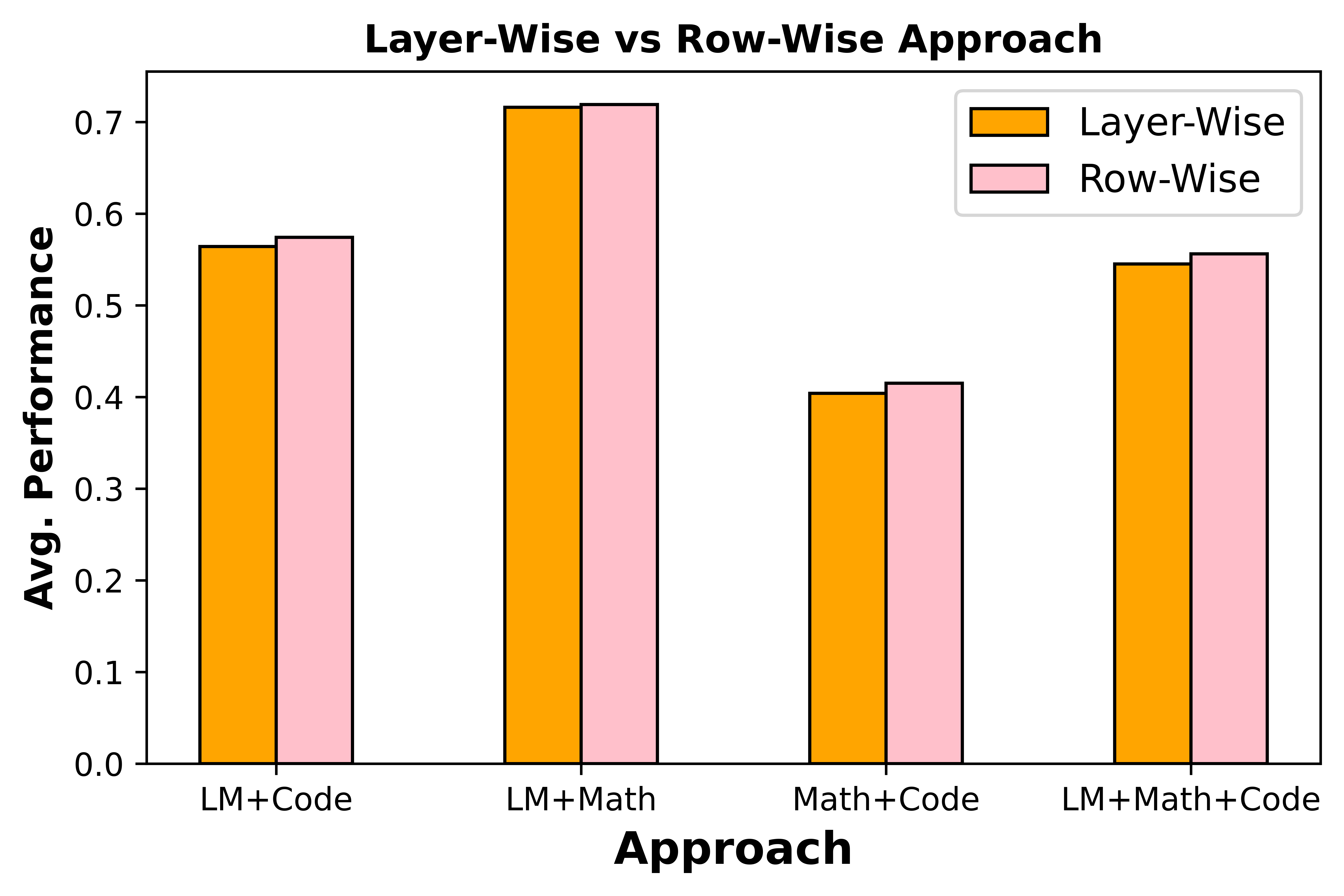}
    \caption{Layer-Wise vs Row-Wise Ranking}
    \label{fig:LayervsRow}
\end{figure}

\paragraph{Deciding pruning groups.}
For the ranking in Drop Step 1, we experimented with two different types of ranking: row-wise and layer-wise. In row-wise ranking, we rank each delta parameter of a node (rows of the matrix layer matrix), similarly for layer-wise, we rank it within the entire layer. \Cref{fig:LayervsRow} shows that row-wise ranking achieves a higher average score than the layer-wise approach across the 4 merge combinations using \approach$_{p \propto r}$.

\begin{table}
\centering
\resizebox{0.5\textwidth}{!}{%
\begin{tabular}{lllcccc}
\hline
\textbf{Merging} &
  \multicolumn{1}{c}{\textbf{Models}} &
  \multicolumn{1}{c}{\textbf{Method}} &
  \textbf{AlpacaEval} &
  \textbf{GSM8K} &
  \textbf{MBPP} &
  \textbf{Avg.} \\ \hline
\multirow{12}{*}{\begin{tabular}[c]{@{}l@{}}Drop-\\ Elect-\\ Merge\end{tabular}} &
  \multirow{3}{*}{\begin{tabular}[c]{@{}l@{}}LM + \\ Math\end{tabular}} &
  \begin{tabular}[c]{@{}l@{}}\wanda{}$_{p=0/1}$\end{tabular} &
  78.2 &
  63.4 &
  \multicolumn{1}{c|}{/} &
  70.8 \\ \cline{3-7}
 &
   &
  \begin{tabular}[c]{@{}l@{}} \wanda{}$_{p \propto r}$ \end{tabular} &
  81.3 &
  62.5 &
  \multicolumn{1}{c|}{/} &
  71.9 \\ \cline{3-7} 
 &
   &
  \begin{tabular}[c]{@{}l@{}}\approach$_{p \propto r}$\end{tabular} &
  \textbf{81.8} &
  \textbf{64.1} &
  \multicolumn{1}{c|}{/} &
  \textbf{72.9} \\ \cline{2-7} 
 &
  \multirow{3}{*}{\begin{tabular}[c]{@{}l@{}}LM + \\ Code\end{tabular}} &
  
  \begin{tabular}[c]{@{}l@{}}\wanda{}$_{p=0/1}$\end{tabular} &
  54.8 &
  / &
  \multicolumn{1}{c|}{5} &
  29.9 \\ \cline{3-7}
 &
   &
  \begin{tabular}[c]{@{}l@{}} \wanda{}$_{p \propto r}$ \end{tabular} &
  80.7 &
  / &
  \multicolumn{1}{c|}{30.6} &
  55.6 \\ \cline{3-7} 
 &
   &
  \begin{tabular}[c]{@{}l@{}}\approach$_{p \propto r}$\end{tabular} &
  \textbf{79.7} &
  / &
  \multicolumn{1}{c|}{\textbf{35.2}} &
  \textbf{57.5} \\ \cline{2-7} 
 &
  \multirow{3}{*}{\begin{tabular}[c]{@{}l@{}}Math + \\ Code\end{tabular}} &
  \begin{tabular}[c]{@{}l@{}}\wanda{}$_{p=0/1}$\end{tabular} &
  / &
  58.6 &
  \multicolumn{1}{c|}{24} &
  41.3 \\ \cline{3-7}
 &
   &
  \begin{tabular}[c]{@{}l@{}} \wanda{}$_{p \propto r}$ \end{tabular} &
  / &
  63.8 &
  \multicolumn{1}{c|}{17.8} &
  40.8 \\ \cline{3-7}
 &
   &
  \begin{tabular}[c]{@{}l@{}}\approach$_{p \propto r}$\end{tabular} &
  / &
  \textbf{64.9} &
  \multicolumn{1}{c|}{\textbf{20.0}} &
  \textbf{42.4} \\ \cline{2-7} 
 &
  \multirow{3}{*}{\begin{tabular}[c]{@{}l@{}}LM + \\ Math + \\ Code\end{tabular}}
 &
  \begin{tabular}[c]{@{}l@{}}\wanda{}$_{p=0/1}$\end{tabular} &
  67.6 &
  57.8 &
  \multicolumn{1}{c|}{28} &
  51.2 \\ \cline{3-7}
 &
   &
  \begin{tabular}[c]{@{}l@{}} \wanda{}$_{p \propto r}$ \end{tabular} &
  75.4 &
  60.9 &
  \multicolumn{1}{c|}{28.6} &
  55.0 \\ \cline{3-7} 
 &
   &
  \begin{tabular}[c]{@{}l@{}}\approach$_{p \propto r}$\end{tabular} &
  \textbf{80.4} &
  \textbf{61.8} &
  \multicolumn{1}{c|}{\textbf{31.4}} &
  \textbf{57.9} \\ \hline
\end{tabular}
}
\caption{\approach{} against \wanda{} in Drop-Elect-Fuse Setup.}
\label{tab:wanda_comparison}
\end{table}

\paragraph{Importance of GPT4-as-a-judge for Math tasks.}
We perform prompt-based evaluation by prompting GPT4 with the question, reference solution, and model-generated answer to evaluate the correctness \cite{li2024leveraging}. This is a more comprehensive automatic evaluation as compared to the hard-coded parsing approaches that are suboptimal math evaluators. Refer to \Cref{sec:GPT4Example} for an example showcasing the effectiveness of using GPT4-as-a-judge for evaluating math tasks.

\paragraph{Using activations to rank delta parameters.}
Inspired by \wanda{}, \cite{sun2024a}, we used the product of delta parameter magnitudes and input activations, the higher score of which corresponds to a higher rank of delta parameters in Step 1. The activations are computed for each layer using 128 data samples from GSM8K for the Math-expert \cite{cobbe2021gsm8k}, MBPP train for the Code-expert \cite{mbpp}, and C4 train dataset \cite{Raffel2019ExploringTL} for the LM-expert. \Cref{tab:wanda_comparison} shows the comparison of \approach{} with \magp{} pruning with \wanda{} with \topk{} and \magp{} pruning. We observe the \approach{} outperforms \wanda{} in all the task performances and average scores. We also note that \wanda{} with \magp{} outperforms \wanda{} with \topk{} in three of the four merges but performs 0.5 points worse on the math+code merging.

\section{Related Work}

\textbf{Supervised Fine-tuning (SFT)} is a widely adopted technique in natural language processing, used to equip pre-trained LMs with expert capabilities by optimizing them on task-specific data \cite{dodge2020fine, zhao2023survey}. This paradigm has emerged as the de facto standard approach. Generally, Supervised Fine-tuning (SFT) of LMs can be categorized into two distinct methodologies:
\begin{enumerate}
    \item \textbf{Full Fine-tuning:} All parameters of the pre-trained LM are updated during the fine-tuning process, allowing the model to adapt extensively to the target task or domain \cite{radford2018improving, kenton2019bert}.
    \item \textbf{Parameter-efficient Fine-tuning:} updates only a subset of the pre-trained LM's parameters, thereby preserving the knowledge acquired during the initial pre-training phase while enabling efficient adaptation to the target task or domain \cite{houlsby2019parameter, li2021prefix, hu2021lora, lester2021power, liu2023gpt}. 
\end{enumerate}

\noindent \textbf{Model pruning} techniques are a form of model compression that reduces the number of parameters in a model while maintaining the model's performance  \cite{zhu2017prune, liu2018rethinking, frankle2018the, zhu2023compression}. Pruning methods employ various strategies to select parameters to retain, including using magnitude-based pruning \cite{ Li2018OptimizationBL, lee2021layeradaptive}, leveraging the Hessian matrix \cite{pmlr-v202-frantar23a}, combining weight magnitudes with input activations \cite{sun2024a}, minimizing discrepancies between source and pruned model parameters \cite{zhang2023dynamic}, and introducing nonuniform layered sparsity for higher pruning rates \cite{yin2023outlier}. While these pruning techniques share the goal of parameter reduction, there is a notable difference compared to \approach{}. \approach{} focuses on pruning delta parameters instead of fine-tuned model parameters.\\

\noindent \textbf{Model merging} techniques focus on combining multiple task-specific models into a single model with diverse capabilities without requiring the original training data \cite{NEURIPS2023_299a08ee}. Widely recognised methods in model merging include Average Merging \cite{wortsman2022model}, which constructs merged models using averaged SFT parameters; Task Arithmetic \cite{ilharco2023editing}, which uses domain-specific offsets and pre-defined scaling terms to distinguish the importance of each model being merged; Fisher Merging \cite{Fisher-merging}, utilizing the Fisher information matrix for weighted parameter fusion \cite{fisher1922mathematical}; RegMean \cite{jin2023dataless}, which optimizes a linear regression problem for model merging; and \ties{}-Merging \cite{ties}, which resolves conflicts by trimming low-magnitude parameters, resolving sign disagreements, and disjointly merging parameters with consistent signs. Recent approaches such as \dare{} \cite{dare} and DPPA \cite{zhu2024dppa} aim to reduce interference during merging by dropping delta parameters and rescaling remaining delta parameters to maintain each model's performance before merging. \\

This paper focuses its contribution on merging homologous models, where model merging is achieved by manipulating delta parameters. The baselines we use in this paper primarily involve manipulating delta parameters of homologous models.

\section{Conclusion}

We introduced a novel approach for merging homologous models, \approach{}. This approach consists of three steps: Drop, Elect and Merge. For the Drop step, We proposed \textbf{\magp{}}, a novel pruning method that samples delta parameters based on their magnitudes. We showed that \approach{} encompasses \nodrop{}, \dare{} and \ties{} and we can cast \approach{} as these methods by adjusting hyperparameters. We found that \approach{} achieves an average improvement of 2.4 points over baseline methods using pruning techniques(\dare{}, \ties{}) and 11.1 points over the method without pruning(\ta{}). \approach{} achieved the highest average score for 5 out of 8 merges and the second-best score for 2 out of 3 remaining merges (\Cref{tab:three_step_results,tab:TA_results}).

\section{Limitations}
\approach{} has more hyperparameters compared to other baselines like \dare{} and \ties{}. While this allows our approach to be more customized to each model merging, it adds additional complexity to finding the optimal merging. Due to computational constraints and a limited number of on-the-shelf models available, we could not extensively evaluate models with different backbones and sizes. Similar to \dare{}, \ties{} and \ta{}, our approach is only effective for models with the same backbone model. Some potential future work could involve exploring applications of these techniques for merging models with different pre-trained backbones.
\section{Potential Risks}
LLMs can be used for harmful content generation and for spreading misinformation.
\section{Ethical Considerations}

Not applicable.
\bibliography{custom}
\clearpage
\appendix

\section{Appendix}
\label{sec:appendix}

\subsection{Importance of GPT4-as-a-judge for Math tasks - Example} \label{sec:GPT4Example}

\begin{quote}

    \textbf{Question:} Each person in a certain household consumes 0.2 kg of rice every meal. Supposing 5 members of the household eat rice every lunch and dinner, how many weeks will a 42 kg bag of rice last?

    \textbf{Generated Answer:} 1. First, we need to determine the total amount of rice consumed by the household in one day. Since each person consumes 0.2 kg of rice per meal and there are 5 people in the household, the total amount of rice consumed in one day is:

       0.2 kg/person/meal * 5 people * 2 meals/day = 2 kg/day

    2. Next, we need to calculate the total amount of rice consumed in a week. There are 7 days in a week, so the total amount of rice consumed in a week is:
    
       2 kg/day * 7 days/week = 14 kg/week
    
    3. Finally, we need to determine how many weeks a 42 kg bag of rice will last. To do this, we will divide the total weight of the bag by the total amount of rice consumed in a week:
    
       42 kg / 14 kg/week = 3 weeks
    
    So, a 42 kg bag of rice will last for 3 weeks for a household of 5 people who eat rice at every meal.

    \textbf{GPT4 Evaluation Comment:} The model's response is correct. It followed the correct process of determining the daily consumption of rice, converting that to weekly consumption, and then determining the number of weeks a 42 kg bag of rice would last under those conditions. The computation aligns correctly with the reference solution.
\end{quote}

Given the example above, a hard-coded parser would take the last number in the generated answer, 5, as the final answer and mistakenly label the answer as incorrect. In contrast, GPT4 acts as a smart parser and correctly identifies the final answer as 3 and labels it as correct. We also note that GPT-4 evaluated the intermediate steps in the solution to ensure their correctness. This demonstrates the limitations of inflexible parsing methods and the benefits of using language models that understand the context for automatic evaluation.

\subsection{Pruning Experiments}
We compare the performance of \approach{} against the \dare{} baseline to show that magnitude sampling improves the selection of delta parameters to retain and better maintain the model's task performance. We vary the drop rate \textit{p} in [0.3, 0.5, 0.7, 0.8, 0.9, 0.91, 0.92, 0.93, 0.94] and apply the \dare{} and \approach{} to get models after removing the proportion of delta parameters. We then evaluate the model's performance on its corresponding SFT task. \Cref{tab:pruningRowvsLayer} shows the comparison between \dare{}, random ranking and \magp{}. We performed experiments with random ranking where we assign ranks to parameters randomly in step1. Comparing \magp{} with random ranking would enable us to see the effectiveness of using delta parameter magnitudes as a ranking method. Overall, \magp{} performs better than \dare{} and random ranking on average by 1.3 points and 1.6 points. Similar to the results from the merging experiment, row-wise \magp{} performs better than layer-wise by 1 point.

\begin{table}[ht]
\centering
\resizebox{0.5\textwidth}{!}{%
\begin{tabular}{c|ccccc}
\hline
Sparsity & \dare{} & 
\begin{tabular}[c]{@{}l@{}}\magp{}\\ (row)\end{tabular} &
\begin{tabular}[c]{@{}l@{}}\magp{}\\ (layer)\end{tabular} &
\begin{tabular}[c]{@{}l@{}}random rank\\ (row)\end{tabular} &
\begin{tabular}[c]{@{}l@{}}random rank\\ (layer)\end{tabular} \\
\hline
0.3 & 0.636 & \textbf{0.652} & 0.642 & 0.630 & 0.642 \\
0.5 & 0.627 & 0.648 & 0.633 & \textbf{0.655} & 0.621 \\
0.7 & 0.615 & \textbf{0.642} & 0.612 & 0.636 & 0.615 \\
0.8 & \textbf{0.658} & 0.645 & 0.624 & 0.655 & 0.621 \\
0.9 & 0.618 & \textbf{0.639} & 0.621 & 0.621 & 0.618 \\
0.91 & 0.633 & 0.618 & \textbf{0.639} & 0.633 & 0.609 \\
0.92 & 0.615 & 0.630 & 0.618 & 0.588 & \textbf{0.652} \\
0.93 & \textbf{0.652} & 0.648 & 0.642 & 0.591 & 0.642 \\
0.94 & 0.597 & \textbf{0.645} & 0.642 & 0.555 & 0.600 \\
\hline
\end{tabular}
}
\caption{Wizardmath-13B-V1.0 Row-Wise vs Layer-Wise ranking Pruning Results}
\label{tab:pruningRowvsLayer}
\end{table}

\subsection{Pruning Rate Hyperparameter Search For Model Merging}

\Cref{tab:PruningRateResults} shows the results of the pruning rate hyperparameter search for each merging combination. While both \magp{} and \dare{} can maintain the performance of individual expert model performance up to a high drop rate of 0.9, our findings indicate that a drop rate of 0.5, works best for LM+Math, Math+Code and LM+Math+Code. For LM+Code, a drop rate of 0.7 is optimal. Thus, we can infer that while dropping delta parameters helps reduce interference during merging, dropping too many parameters may lead to the loss of information useful for effective merging.

\begin{table}[ht]
\resizebox{0.5\textwidth}{!}{%
\begin{tabular}{|l|l|l|c|c|c|c|}
\hline
Models & Drop rate & AlpacaEval & GSM8K & MBPP & Average \\
\hline
\multirow{5}{*}{\begin{tabular}[c]{@{}l@{}}LM + \\ Math\end{tabular}} 
 & 0.1 & 0.805 & 0.599 & / & 0.702\\
 & 0.3 & \textbf{0.812} & 0.629 & / & 0.721\\
 & 0.5 & 0.804 & \textbf{0.645} & / & \textbf{0.724}\\
 & 0.7 & 0.787 & 0.611 & / & 0.699 \\
 & 0.9 & 0.683 & 0.455 & / & 0.570 \\
\hline
\multirow{5}{*}{\begin{tabular}[c]{@{}l@{}}LM + \\ Code\end{tabular}} 
 & 0.1 & 0.741 & / & 0 & 0.370 \\
 & 0.3 & 0.770 & / & 0 & 0.385 \\
 & 0.5 & \textbf{0.802} & / & 0.152 & 0.477 \\
 & 0.7 & 0.798 & / & \textbf{0.34} & \textbf{0.569} \\
 & 0.9 & 0.737 & / & 0.262 & 0.500 \\
\hline
\multirow{5}{*}{\begin{tabular}[c]{@{}l@{}}Math + \\ Code\end{tabular}} 
 & 0.1 & / & 0.619 & 0.166 & 0.393 \\
 & 0.3 &  / & 0.618 & 0.184 & 0.401 \\
 & 0.5 &  / & 0.626 & \textbf{0.206} & \textbf{0.416} \\
 & 0.7 &  / & \textbf{0.633} & 0.19 & 0.412 \\
 & 0.9 &  / & 0.622 & 0.128 & 0.375 \\
\hline
\multirow{5}{*}{\begin{tabular}[c]{@{}l@{}}LM + \\ Math + \\ Code\end{tabular}}
 & 0.1 & 0.732 & 0.545 & 0.114 & 0.464 \\
 & 0.3 & 0.766 & 0.623 & \textbf{0.302} & 0.564\\
 & 0.5 & \textbf{0.794} & \textbf{0.630} & 0.3 & \textbf{0.575} \\
 & 0.7 & 0.770 & 0.622 & 0.23 & 0.541 \\
 & 0.9 & 0.688 & 0.446 & 0.128 & 0.421 \\
\hline
\end{tabular}
}
\caption{Drop Rate of parameters against Task performance}
\label{tab:PruningRateResults}
\end{table}


%
%
%
%
%
%
%
%
\end{document}